\documentclass[10pt, a4paper]{article}
\usepackage{graphicx}
\usepackage{enumerate}
\usepackage{subfigure}
\usepackage{microtype}

\usepackage[]{lrec-coling2024} 
\title{Hypothesis Testing Prompting Improves Deductive Reasoning in Large Language Models}

\name{Yitian Li$^{1,2}$, Jidong Tian$^{1,2}$, Hao He$^{1,2}$, Yaohui Jin$^{1,2}$} 

\address{$^1$MoE Key Lab of Artificial Intelligence, AI Institute, Shanghai Jiao Tong University \\$^2$State Key Lab of Advanced Optical Communication System and Network\\
         \{yitian\_li, frank92, hehao, jinyh\}@sjtu.edu.cn\\}

\abstract{
Combining different forms of prompts with pre-trained large language models has yielded remarkable results on reasoning tasks (e.g. Chain-of-Thought prompting). However, along with testing on more complex reasoning, these methods also expose problems such as invalid reasoning and fictional reasoning paths. In this paper, we develop \textit{Hypothesis Testing Prompting}, which adds conclusion assumptions, backward reasoning, and fact verification during intermediate reasoning steps. \textit{Hypothesis Testing prompting} involves multiple assumptions and reverses validation of conclusions leading to its unique correct answer. Experiments on two challenging deductive reasoning datasets ProofWriter and RuleTaker show that hypothesis testing prompting not only significantly improves the effect, but also generates a more reasonable and standardized reasoning process.
 \\ \newline \Keywords{Deductive Reasoning, Large Language Models, Prompt} }

\begin{document}

\maketitleabstract

\section{Introduction}

The release of large language models (LLMs) has revolutionized the NLP landscape recently~\cite{DBLP:journals/corr/abs-2201-08239,DBLP:journals/corr/abs-2001-08361,DBLP:journals/corr/abs-2204-02311}. Scaling up the size of language models and conducting diversified prompt methods become mainstream~\cite{DBLP:journals/csur/LiuYFJHN23,DBLP:journals/tmlr/WeiTBRZBYBZMCHVLDF22,DBLP:journals/corr/abs-2304-13712}. Given In-context learning or Chain-of-Thought prompts have already achieved high performance on challenging tasks such as commonsense, arithmetic, and symbolic reasoning~\cite{DBLP:journals/corr/abs-2303-05398,DBLP:conf/emnlp/00010O21,DBLP:conf/nips/KojimaGRMI22}. Logical reasoning is one of the most important and long-standing problems in NLP~\cite{doi:10.1126/science.aaa8685,DBLP:books/daglib/0023820}, and integrating this ability into natural language understanding systems has always been a goal pursued~\cite{DBLP:conf/icml/DuLTM22}. 

\begin{figure}[!tb]
\centering
\includegraphics[width=8.0cm]{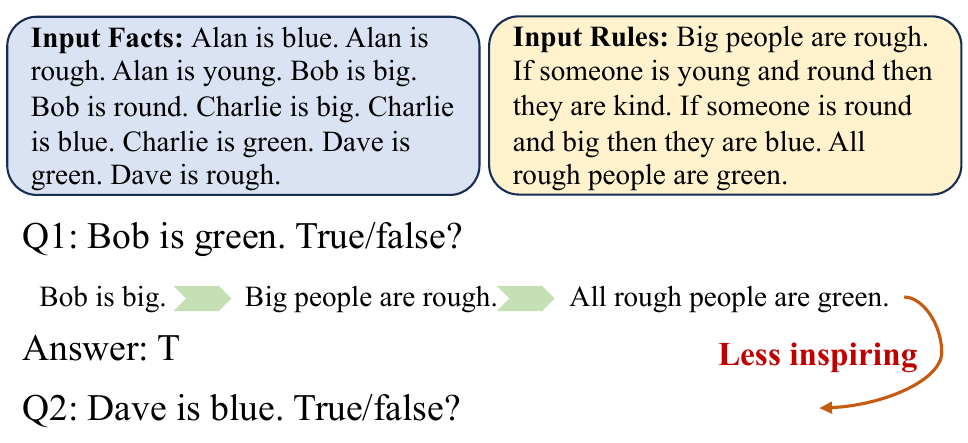}
\caption{Questions in RuleTaker involve logical reasoning with facts and rules.}
\label{example}
\end{figure}

Nevertheless, scaling has been demonstrated to offer limited advantages in resolving complex logical reasoning issues~\cite{DBLP:journals/corr/abs-2212-13894}. For example, \citet{DBLP:journals/corr/abs-2210-01240} show that Chain-of-Thought prompting struggles with proof planning for more complex logical reasoning problems. Additionally, the performance suffers greatly while handling recently released and out-of-distribution logical reasoning datasets~\cite{DBLP:journals/corr/abs-2304-03439}. Despite many works have explored variants of Chain-of-Thought prompts to facilitate LLMs inference~\cite{DBLP:conf/nips/ZelikmanWMG22,zheng2023progressivehint}, we discover that the present logical reasoning task prompts place an excessive amount of emphasis on the reasoning process while ignoring the origin, purpose, and effectiveness of reasoning~\cite{DBLP:journals/corr/abs-2205-09712,xi2023selfpolish}. As examples shown in Figure~\ref{example}, the difficulty in judging logical problems arises not only from the process of reasoning but also from the choice of facts and rules to use as a starting point. Even if we were provided the thought process for some of the issues, it would not be very beneficial for others, based on how we previously created the prompts.    

In this paper, we propose \textit{Hypothesis Testing Prompting}, a new and more considerate prompt template design idea. Hypothesis testing is a formal procedure for investigating our ideas about the world using statistics and is often used by scientists to test specific predictions~\cite{bevans-2022}. We draw inspiration from its process to introduce a process of conclusion assumptions, backward reasoning, and fact verification. Experiments on RuleTaker~\cite{DBLP:conf/ijcai/ClarkTR20} and ProofWriter~\cite{DBLP:conf/acl/TafjordDC21} show the effectiveness of our novel prompting paradigm as a strategy for promoting deductive reasoning in large language models. Further analyses show that \textit{Hypothesis Test prompting} generates more desirable intermediate processes and significantly improves the "Unknown" label.

\section{Related Work}
\subsection{Few-Shot Prompting}
\citet{DBLP:conf/nips/BrownMRSKDNSSAA20} propose in-context learning as an alternative few-shot prompting way to stimulate ability. Besides, chain-of-Thought (CoT)~\cite{DBLP:conf/nips/Wei0SBIXCLZ22} is one of the most well-known works, which decomposes the problem into intermediate steps and further improves the ability of large language models. Subsequently, several follow-up works were carried out, including Zero-shot-CoT (simply adding "Let's think step by step" before each answer)~\cite{DBLP:conf/nips/KojimaGRMI22}, Self-consistency~\cite{DBLP:journals/corr/abs-2203-11171}, complexity-based~\cite{DBLP:journals/corr/abs-2210-00720}, and other prompting work~\cite{DBLP:journals/corr/abs-2305-12147, DBLP:conf/emnlp/JungQWBB0C22, DBLP:journals/corr/abs-2205-10625,DBLP:journals/corr/abs-2210-01240}. While these methods enhance the performance of inference by paying attention to indications of the reasoning process, they often overlook some aspects such as identifying the root cause of the problem, establishing efficient reasoning strategies, and determining the direction of logical reasoning.
\subsection{Deductive Reasoning}
Deductive reasoning is defined as the application of general concepts to particular circumstances~\cite{johnson2010deductive}. Making logical assumptions is the foundation of deductive reasoning, which then bases a conclusion on those assumptions. The deduction task is then applied to a situation from the actual world after starting with a rule. In light of the principles \textit{"All men are mortal."} and \textit{"Socrates is a man."} for example, we can draw the conclusion that "Socrates is mortal."~\cite{johnson1999deductive}.

\section{Hypothesis Testing Prompting}

Hypothesis testing is a formal procedure for investigating our ideas about the world using statistics and used by scientists to test specific predictions that arise from theories~\cite{bevans-2022,2012Hypothesis}. There are 5 main steps in hypothesis testing:
\begin{enumerate}
	\item State your research hypothesis;
	\item Collect data in a way designed to test the hypothesis;
	\item Perform an appropriate statistical test;
        \item Decide whether to reject or fail to reject your null hypothesis;
        \item Present the findings in your results and discussion section;
\end{enumerate}

\begin{figure*}[t]
\begin{center}
\includegraphics[width=1\textwidth,height=0.95\textwidth]{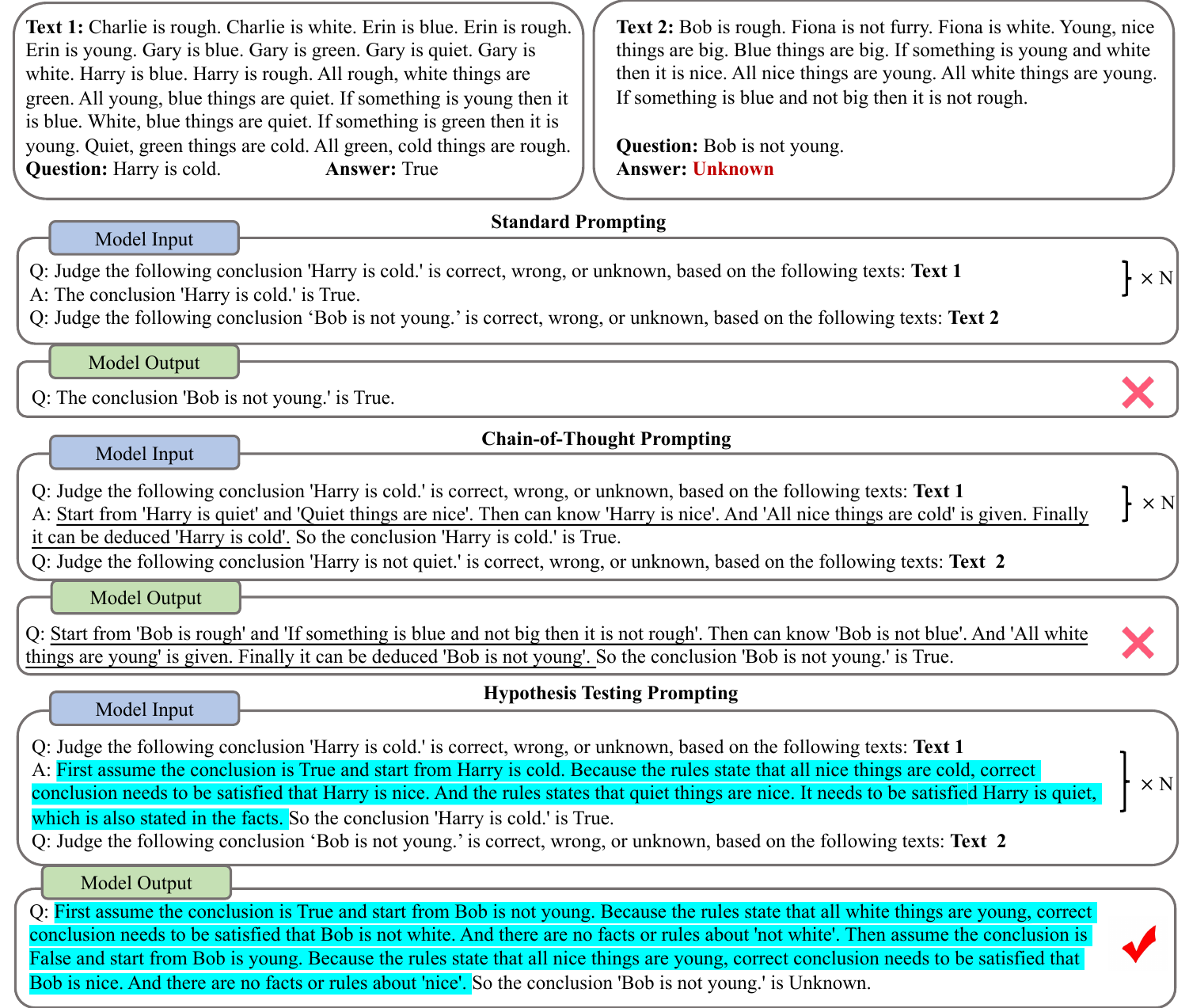}
\end{center}
\caption{Comparison of three prompting methods: (a) Standard (b) Chain-of-Thought (c) Hypothesis Testing. Particularly, we highlight the Hypothesis testing reasoning processes. The comparative experimental results show that: Hypothesis testing prompting enables large language models to tackle complex logical reasoning. }
\label{fig:d}
\end{figure*}

When completing a challenging reasoning activity, such as a multi-step deductive reasoning problem, one is not conducting random reasoning to obtain all possible intermediate results. We shall choose the relevant conditions for inference verification after initially making assumptions about the judgment problem, such as \textit{" First assume the conclusion is True and start from ... Then assume the conclusion is False and start from ... because the rules state that ... So the conclusion ..."}. The purpose of this study is to give language models the capacity to build a process that is similar to what we defined as \textit{Hypothesis Testing Prompting}. We will show that large language models can generate more appropriate thought and more accurate results if demonstrations of hypothesis test prompting are provided in the exemplars for few-shot prompting. Figure~\ref{fig:d} shows an example of a model producing a hypothesis testing thought to solve a deductive reasoning problem.

\section{Experiment}
\subsection{Experimental Setup}
We explore \textit{Hypothesis Test Prompting} for ChatGPT (GPT-3.5-Turbo in the OpenAI API) on multiple logical reasoning benchmarks.

\textbf{Benchmarks.} 
Considering FOL reasoning in question answering systems, there are two world assumptions~\cite{REITER1981119} that result in different objectives. One is the \textbf{closed world assumption} (CWA), which is the presumption that what is not currently known to be entailment is contradiction. The other is the \textbf{open world assumption} (OWA), whose objective should distinguish false propositions from uncertain ones. Due to differences in world assumptions, our analysis and solutions are also different.

We consider the following two deductive reasoning problem benchmarks: (1) the RuleTaker~\cite{DBLP:conf/ijcai/ClarkTR20} benchmark using CWA assumption; (2) the ProofWriter~\cite{DBLP:conf/acl/TafjordDC21} benchmark using OWA assumption. Both datasets are divided into five parts, each part requiring 0, $\leq$ 1, $\leq$ 2, $\leq$ 3, and $\leq$ 5 hops of reasoning, respectively. We conducted comparison tests on the test set of the two datasets for 5 distinct hops.

\textbf{Standard prompting.} As one of the baselines, we take into account the common few-shot prompting, made popular by \citet{DBLP:conf/nips/BrownMRSKDNSSAA20}, in which a language model is provided with in-context examples of input-output pairings before producing a prediction for a test-time example. Examples are presented in the form of questions and answers. As seen in Figure~\ref{fig:d}(above), the model directly answers the question.

\textbf{Chain-of-Thought prompting.} We also compare with Chain-of-thought prompting which has achieved encouraging results on complex reasoning tasks~\cite{DBLP:conf/nips/Wei0SBIXCLZ22}. As seen in Figure~\ref{fig:d}(middle), the model not only provides the final answer but also comes with the consideration of intermediate steps.

\begin{figure*}[htbp]
\centering 
\subfigure[]{   
\begin{minipage}{0.48\linewidth}
\centering   
\includegraphics[width=1\linewidth]{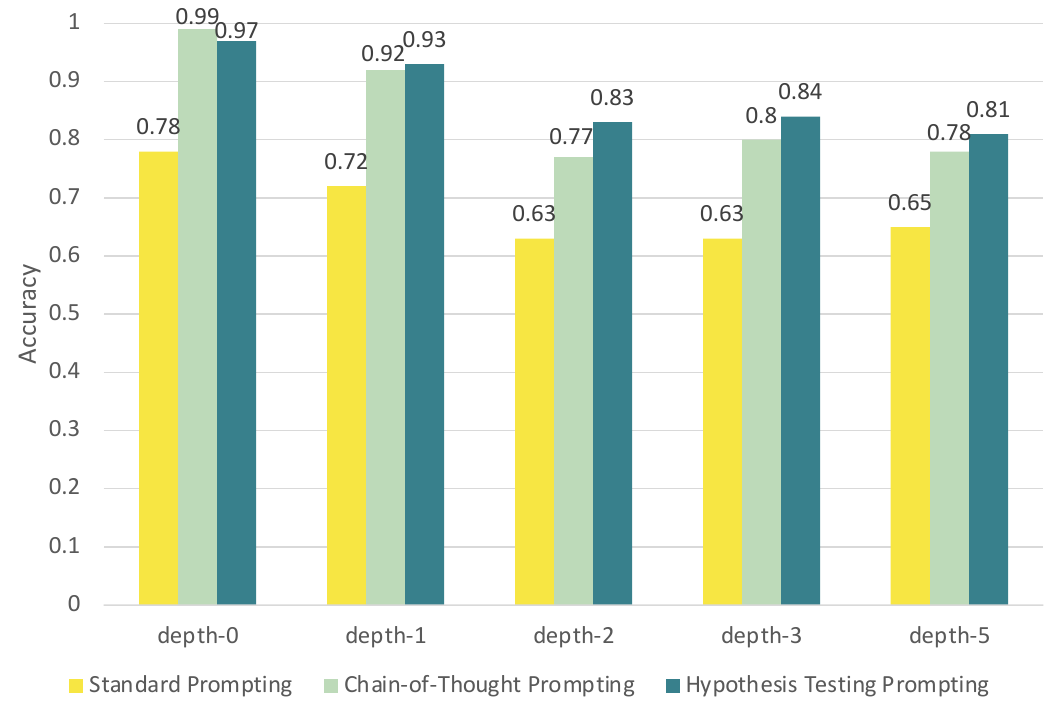}  
\end{minipage}
}
\subfigure[]{ 
\begin{minipage}{0.48\linewidth}
\centering    
\includegraphics[width=1\linewidth]{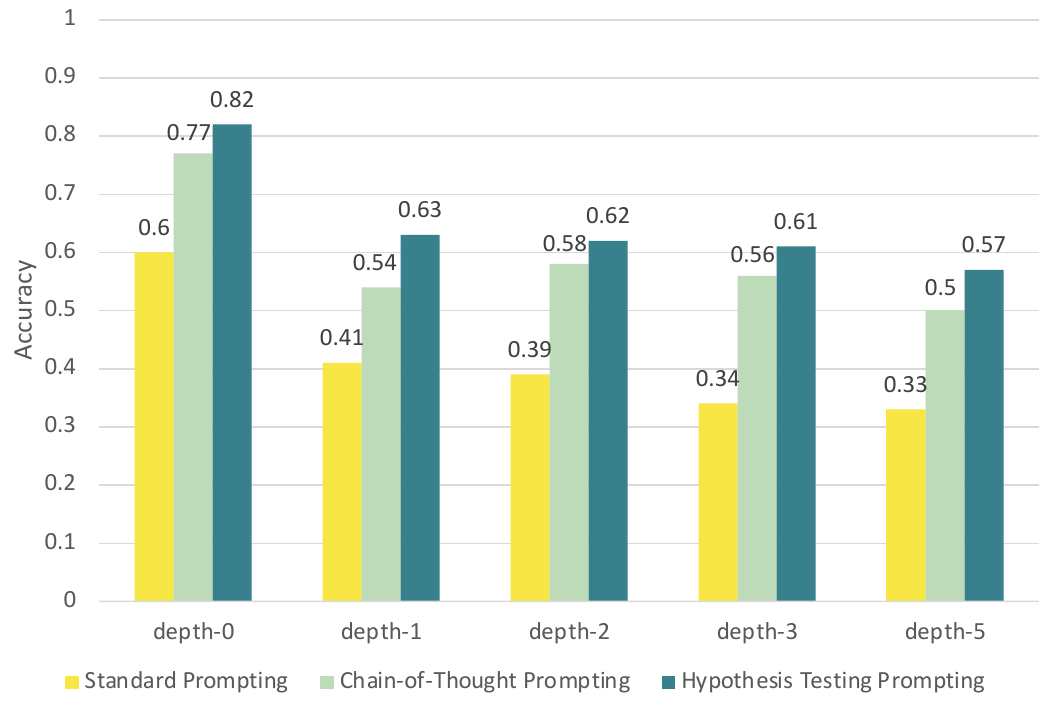}
\end{minipage}
}
\caption{Prediction accuracy results on the (a) RuleTaker and (b) ProofWriter datasets.}  
\label{result1}    
\end{figure*}

\textbf{Hypothesis Testing Prompting.} Our proposed approach is to augment each exemplars in few-shot prompting with the thought of hypothesis testing for an associated answer, as illustrated in Figure~\ref{fig:d}(below). We show one chain of thought exemplars (\textit{Example}: \textit{Judge the following conclusion '<Conclusion>' is true, false, or unknown, based on the following facts and rules: <Facts> ... <Rules> ...}).

\subsection{Experimental Results}

The results for Hypothesis Testing Prompting and the baselines on the RuleTaker datasets are provided in Figure~\ref{result1}(a), and ProofWriter results are shown in Figure~\ref{result1}(b). From the results, we observe that our method significantly outperforms the other two baselines, especially on ProofWriter. Figure~\ref{result1}(a) demonstrates that while CoT performs well in the low hop, Hypothesis Testing prompting performs better as the hops count increases on RuleTaker. While on ProofWriter, our approach has a thorough lead (improved accuracy by over 4\% on all hops). 
Comparing two datasets, the latter distinguishes between "False" and "Unknown", which demand a greater level of logic. The results on two datasets that were analyzed show a weakness in all methods for handling "Unknown" labels. This beacuse the OWA hypothesis necessitates the exclusion of both positive and negative findings to validate the "Unknown" label. The advantages of our strategy are illustrated by the comparison of the model output outputs in Figure~\ref{fig:d}. The content \textit{"First assume the conclusion is True ... Then assume the conclusion is False ... So ... is Unknown."} generated by the model through learning Hypothesis Testing prompting is more in line with our thinking. Besides, we'll conduct further research and show it later.

\begin{figure}[t]
\centering
\subfigure[The proof accuracy of Chain-of-Thought and Hypothesis Testing prompting.]{
\begin{minipage}[b]{0.45\textwidth}
\includegraphics[width=1\textwidth]{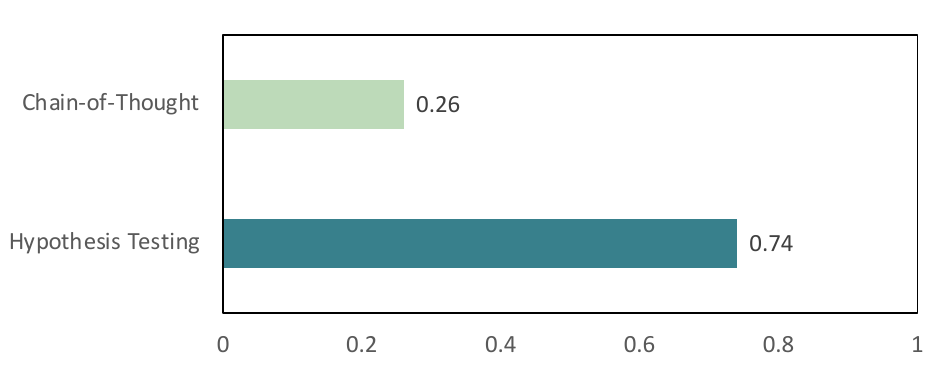}
\end{minipage}}
\subfigure[Comparison of accuracy between Chain-of-Thought and Hypothesis Testing prompting on "Unknown" label. ]{
\begin{minipage}[b]{0.45\textwidth}
\includegraphics[width=1\textwidth]{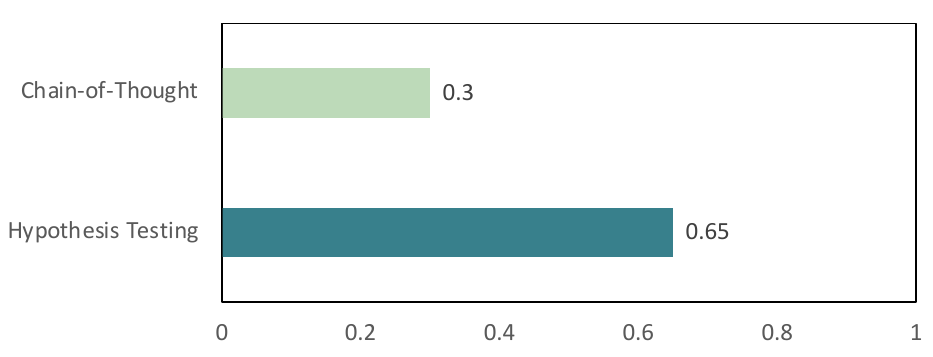}
\end{minipage}}
\caption{Further results on ProofWriter. } 
\label{case_2}
\end{figure} 
\subsection{Further Analysis}

We carry out the following thorough analysis to better comprehend the thought process:

\textbf{Proof Accuracy.}
Five students are required to manually evaluate the outcomes of the intermediate reasoning after we randomly picked 100 examples from depth-5 of the ProofWriter. Proof accuracy represents the proportion where the inference process has been proven to be reasonable in the correct part of data label prediction. We compare the results of Chian-of-Thought and Hypothesis Testing prompting and report in Figure~\ref{case_2}(a). While Hypothesis Testing prompting mostly produced the correct intermediate reasoning process when the predicted label was correct, CoT only generated the correct chain for 26\% of the examples. This result is in line with other research showing that LMs rely on spurious correlations when solving logical problems from beginning to end. Additionally, our approach can successfully increase reasoning's rationality. In processing the "Unknown" label, Hypothesis Testing prompting performs noticeably better than Chain-of-Thought.

\textbf{"Unknown" accuracy.}
In the ProofWriter dataset, we separately counted the accuracy of the "Unknown" label shown in Figure~\ref{case_2}(b). The results point to a flaw in the Chain-of-Thought strategy's handling of "Unknown" labels(only 0.3 accuracyies). Contrarily, Hypothesis Testing prompting significantly increases the reliability of judging this label (up to 0.65). This further illustrates the value of holding various assumptions, as well as the reverse confirmation of conclusions.

\section{Conclusion}

We have investigated Hypothesis Testing prompting as a straightforward and widely applicable technique for improving deductive reasoning in large language models. Multiple assumptions are made during hypothesis testing, and conclusions are reverse-validated to arrive at the one and only accurate answer. Through experiments on two logical reasoning datasets, we find that Hypothesis Testing prompting allows large language models to construct reasoning more reasonably and accurately. We anticipate that additional research on language-based reasoning approaches will be stimulated by our novel prompting design strategy.


\nocite{*}
\section{References}\label{sec:reference}

\bibliographystyle{lrec-coling2024-natbib}
\bibliography{lrec-coling2024-example}


\end{document}